\documentclass[11pt]{article}
\usepackage[preprint]{acl}

\usepackage{times}
\usepackage{latexsym}
\usepackage{amsmath}
\usepackage{amssymb}
\usepackage{booktabs}
\usepackage{graphicx}
\usepackage{tcolorbox}
\usepackage{multirow}
\usepackage{booktabs}
\usepackage{xcolor}
\usepackage{colortbl}
\usepackage{siunitx}
\usepackage{makecell}
\usepackage[T1]{fontenc}
\usepackage[utf8]{inputenc}
\usepackage{microtype}
\usepackage{inconsolata}
\usepackage{graphicx}

\title{MUSE: Multi-Domain Chinese User Simulation via Self-Evolving Profiles and Rubric-Guided Alignment}

\author{
  \textbf{Zihao Liu}$^{1,2}$\thanks{\ \ Equal contribution.}\ \thanks{\ \ Work done during an internship at Meituan.} \quad
  \textbf{Hantao Zhou}$^2$\footnotemark[1] \quad
  \textbf{Jiguo Li}$^2$ \quad
  \textbf{Jun Xu}$^2$ \quad
  \textbf{Jiuchong Gao}$^2$ \\
  \textbf{Jinghua Hao}$^2$ \quad
  \textbf{Renqing He}$^2$ \quad
  \textbf{Peng Wang}$^1$\thanks{\ \ Corresponding author.} \\
  $^1$College of Computer Science and Artificial Intelligence, Fudan University \\
  $^2$Meituan \\
  \texttt{zihaoliu24@m.fudan.edu.cn}, \texttt{zhouhantao@meituan.com} \\
  \texttt{pengwang5@fudan.edu.cn}
}

\begin{document}
\maketitle

\begin{abstract}
User simulators are essential for the scalable training and evaluation of interactive AI systems. However, existing approaches often rely on shallow user profiling, struggle to maintain persona consistency over long interactions, and are largely limited to English or single-domain settings. We present MUSE, a multi-domain Chinese user simulation framework designed to generate human-like, controllable, and behaviorally consistent responses.
First, we propose Iterative Profile Self-Evolution (IPSE), which gradually optimizes user profiles by comparing and reasoning discrepancies between simulated trajectories and real dialogue behaviors.
We then apply Role-Reversal Supervised Fine-Tuning to improve local response realism and human-like expression. To enable fine-grained behavioral alignment, we further train a specialized rubric-based reward model and incorporate it into rubric-guided multi-turn reinforcement learning, which optimizes the simulator at the dialogue level and enhances long-horizon behavioral consistency. Experiments show that MUSE consistently outperforms strong baselines in both utterance-level and session-level evaluations, generating responses that are more realistic, coherent, and persona-consistent over extended interactions.
\end{abstract}

\section{Introduction}
The rapid evolution of Large Language Models (LLMs) has revolutionized the landscape of conversational AI, enabling agents to handle increasingly complex tasks ranging from open-domain chitchat to professional consultation \cite{ouyang2022traininglanguagemodelsfollow,zhao2024wildchat1mchatgptinteraction}. 
As interactive AI systems become more sophisticated, traditional human-in-the-loop paradigms for building and evaluating them have become prohibitively expensive and unscalable. 
Consequently, User Simulation—leveraging LLMs to mimic human behavior—has emerged as a pivotal solution, enabling progress in such domains as agentic reinforcement learning~\cite{qian2025userrltraininginteractiveusercentric,zhao2025muarlmultiturnuserinteractingagent}, synthetic data generation~\cite{zhu2025dialogueforgellmsimulationhumanchatbot,abbasiantaeb2023letllmstalksimulating}, and multi-turn evaluation~\cite{yao2024taubenchbenchmarktoolagentuserinteraction,barres2025tau2benchevaluatingconversationalagents,Yuan_Chen_Liu_Li_Tang_Zhang_Wang_Wang_Liu_2025} in a scalable and controllable manner.

Despite their promise, existing approaches face three critical limitations. First, current methods for user profile construction predominantly rely on either hand-crafted rules~\cite{wang2025knowbettermodelinghumanlike} or simplistic one-pass LLM extraction~\cite{zhu2025dialogueforgellmsimulationhumanchatbot}. Rule-based approaches require labor-intensive expertise and scale poorly, while naive LLM extraction yields flattened personas that fail to capture nuanced behavioral characteristics. 
Second, maintaining long-term persona consistency and contextual coherence across multi-turn dialogues remains a significant challenge~\cite{choi2025examiningidentitydriftconversations}. As interactions proceed, simulators gradually deviate from assigned identities or abandon role-appropriate behavior over extended interactions.
Third, existing user simulators are predominantly confined to English contexts, with limited exploration in non-English languages such as Chinese.

To bridge these gaps, we present \textbf{MUSE}
(\textbf{M}ulti-Domain Chinese \textbf{U}ser \textbf{S}imulation via self-\textbf{E}volving Profiles and Rubric-Guided Alignment), 
a robust user simulator designed to generate high-fidelity and consistent user behaviors.

We first construct a comprehensive Chinese dialogue dataset spanning six distinct categories, including General Chat, Legal, Medical, and Customer Service. 
Crucially, unlike simplistic one-pass extraction methods that yield sparse descriptions, we propose an \textbf{Iterative Profile Self-Evolution (IPSE)} method. 
Specifically, a profile-conditioned LLM first simulates interactions with an assistant. A reasoning model then evaluates the simulated trajectory against the authentic dialogue to diagnose behavioral deviations and profile deficiencies. Through this iterative generation-and-critique process, IPSE progressively refines the persona, yielding nuanced representations that capture complex user dynamics.

To address the behavioral alignment challenge, we implement a \textbf{Rubric-Guided Reinforcement Learning} framework. 
First, we first train a rubric-based reward model with a two-stage progressive paradigm to obtain a robust evaluator.
Specifically, we begin with Supervised Fine-Tuning (SFT) to jointly align the reward model's Chain-of-Thought (CoT) reasoning and scoring standards with expert annotations. 
This is followed by Reinforcement Learning with Verifiable Rewards (RLVR) using a distance-aware reward mechanism, ensuring the model provides graded feedback.
Equipped with this evaluator, we conduct a Rubric-Guided Multi-turn Reinforcement Learning stage specially on mined hard samples, which effectively improves the model's overall simulation robustness.

Experimental results demonstrate that MUSE consistently outperforms strong baselines in both utterance-level and session-level evaluations, showing its ability to generate user behaviors that are not only more human-like and persona-consistent at the local level, but also more coherent and controllable over extended interactions. These findings highlight the importance of jointly modeling profile quality, local response realism, and long-horizon interaction consistency for realistic user simulation.
Our contributions are summarized as follows:
\begin{itemize} 
    \item We propose MUSE, a robust multi-domain Chinese user simulator. We introduce Iterative Profile Self-Evolution (IPSE), a mechanism that synthesizes nuanced user profiles from raw dialogue histories through iterative refinement, thereby ensuring high consistency in long-term interactions.
    \item We propose a multi-stage training paradigm that employs Role-Reversal SFT for expression alignment and rubric-guided multi-turn RL for behavioral consistency, with an expert-calibrated CoT reward model providing fine-grained, interpretable feedback.
    \item Extensive experiments across six domains demonstrate that MUSE significantly outperforms strong baselines, indicating a substantial advancement in authentic user simulation. 
\end{itemize}

\begin{figure*}[t]
    \centering
    \includegraphics[width=\textwidth]{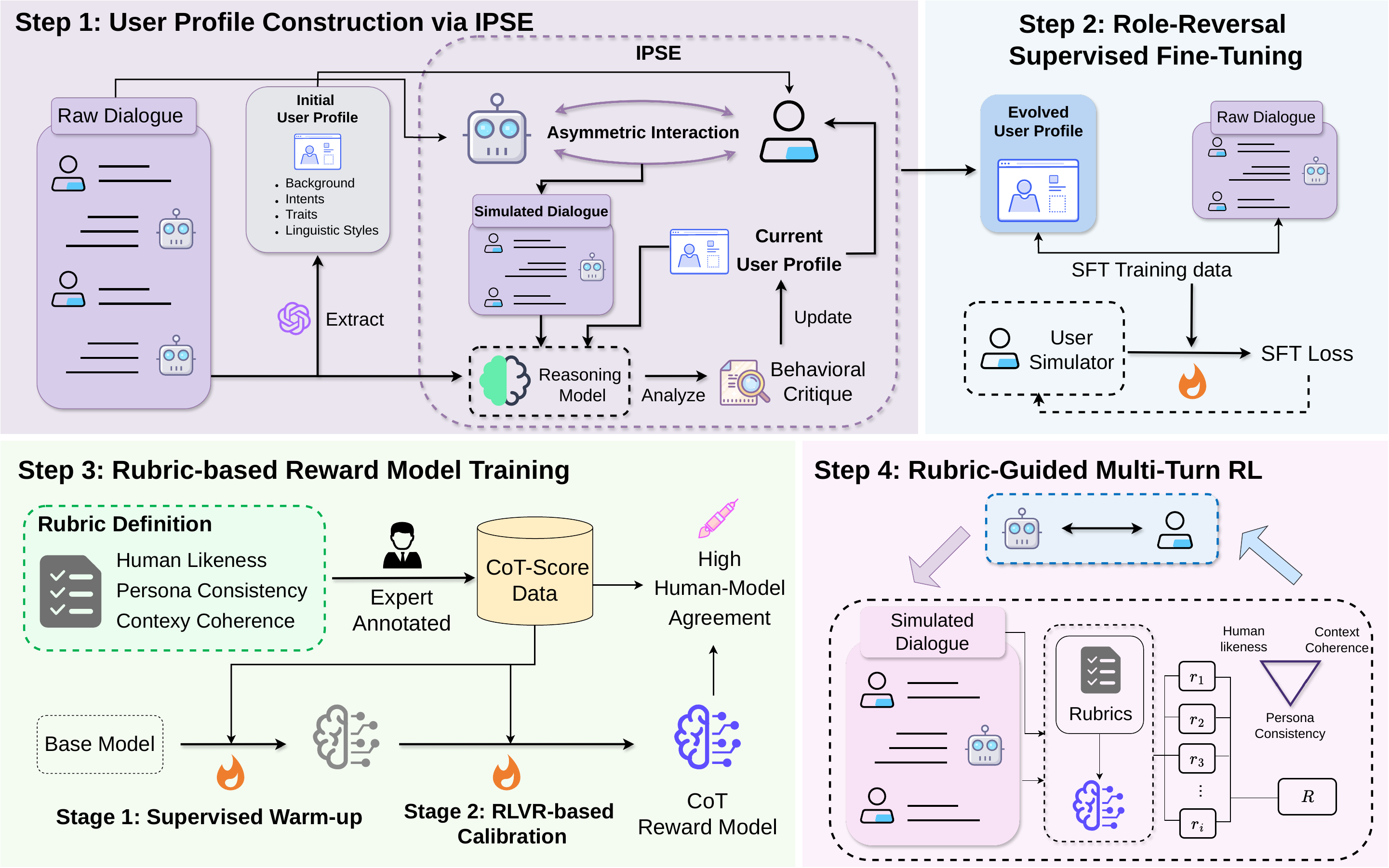}
    \caption{MUSE training pipeline overview. We first employ Iterative Profile Self-Evolution to extract persona-consistent profiles from real dialogues, then apply Role-Reversal SFT for human expression alignment. A two-stage rubric-based reward model is trained for fine-grained feedback, followed by Rubric-Guided multi-turn RL to optimize behavioral alignment in multi-turn interactions.}
    \label{fig:method}
\end{figure*}

\section{Related Work}
\subsection{Traditional User Simulators} 
Traditional user simulators generally fall into rule-based and data-driven categories. The Agenda-Based User Simulator (ABUS)~\citep{schatzmann-etal-2007-agenda} is the most prominent rule-based framework, which utilizes stack-based mechanisms and hand-crafted rules to manage user goals. Although controllable, constructing these rules is labor-intensive and difficult to scale to new domains~\cite{li2017usersimulatortaskcompletiondialogues}. Subsequent data-driven methods~\cite{asri2016sequencetosequencemodelusersimulation, kreyssig-etal-2018-neural} learn policies from dialogue corpora using supervised or reinforcement learning to reduce manual effort. However, these approaches typically rely on fixed dialogue workflow, which limits their generalization capabilities across different domains and restricts the naturalness of generated responses.

\subsection{LLMs as User Simulators} 
The emergence of LLMs has revolutionized user simulation by leveraging their inherent role-playing and zero-shot capabilities. Current approaches generally fall into two categories:

\paragraph{Prompting-based Methods.} 
The most straightforward approach involves instructing off-the-shelf LLMs (\textit{e.g.}, GPT-4) via prompt engineering to adopt a user persona. 
However, recent findings suggest that prompting alone is insufficient for high-fidelity simulation~\cite{naous2026flippingdialoguetrainingevaluating, wang2025knowbettermodelinghumanlike, sekulic-etal-2024-reliable}, arguing that RLHF-tuned models fundamentally differ from human users. They tend to be overly polite, verbose and structurally perfect, failing to capture the messiness of real human typing.

\paragraph{Fine-tuning-based methods.} 
To address the limitations of prompting, recent works explore SFT or RL for better user simulation. \citet{abdulhai2025consistentlysimulatinghumanpersonas} use Multi-Turn RL to penalize persona drift, showing that optimizing for consistency outperforms prompting. \citet{naous2026flippingdialoguetrainingevaluating} train "User LMs" on human data to reduce assistant-like biases. \citet{wang2025knowbettermodelinghumanlike} propose USP with conditional SFT and cycle-consistent RL. However, these methods lack fine-grained RL alignment and are confined to single domains, limiting applicability.

\subsection{Applications of User Simulators}
User simulators primarily serve three major roles: First, they facilitate synthetic data generation, addressing data scarcity and privacy issues while enriching SFT corpora with diverse behaviors~\citep{acikgoz2025desideratumconversationalagentscapabilities, shim2025tooldialmultiturndialoguegeneration}. Second, they provide interactive RL environments, enabling agents to learn robust policies safely without exposing untrained models to real users~\citep{gur2018usermodelingtaskoriented, liu-etal-2023-one}. Thirdly, they interact with agents to assess functional performance metrics, allowing reproducible evaluation across diverse scenarios and user personas~\citep{davidson2023usersimulationlargelanguage, sun2023metaphoricalusersimulatorsevaluating, kazi2024largelanguagemodelsuseragents}.

\section{Methodology}
Our method comprises four phases (Figure~\ref{fig:method}). First, we introduce the \textit{Iterative Profile Self-Evolution} (IPSE) framework to synthesize persona-consistent user profiles across various domains (Section~\ref{sec:ipse}). Second, we perform \textit{Role-Reversal Supervised Fine-Tuning} to align the simulator with authentic human behavior and expression(Section~\ref{sec:rrsft}). Third, we train a specialized \textit{rubric-based reward model} to assess generation quality along three dimensions (Section~\ref{reward_model}). Finally, we use this reward model in \textit{Rubric-Guided Multi-turn Reinforcement Learning} to further align the simulator with fine-grained behavioral standards (Section~\ref{sec:mtrl}).

\subsection{Iterative Profile Self-Evolution}
\label{sec:ipse}
Deriving effective user profiles from real-world conversations is challenging~\cite{wang2025knowbettermodelinghumanlike}. We observe that one-pass extraction with LLMs often yields incomplete or overly generic profiles, which fail to reproduce the distinctive behavioral patterns of real users. 
To address this issue, we propose the \textit{Iterative Profile Self-Evolution} (IPSE) framework, which refines user profiles through iterative dialogue reconstruction and discrepancy-aware reasoning.

\paragraph{Initialization and Formulation}
Given a real dialogue session
\begin{equation}
\mathcal{D}=\{(u_t,a_t)\}_{t=1}^{T},
\end{equation}
where $u_t$ and $a_t$ denote the user and assistant utterances at turn $t$, respectively, our goal is to derive an optimal user profile $\mathcal{P}^*$, instantiated as a system prompt, such that the reconstructed dialogue under $\mathcal{P}^*$ matches the original interaction:
\begin{equation}
\mathcal{P}^*=\arg\min_{\mathcal{P}} \mathrm{Dist}\big(\mathcal{D}, \hat{\mathcal{D}}(\mathcal{P})\big),
\end{equation}
where $\hat{\mathcal{D}}(\mathcal{P})$ is the simulated dialogue induced by profile $\mathcal{P}$. Starting from an initial profile $\mathcal{P}_0$ extracted from $\mathcal{D}$, IPSE iteratively refines the profile through dialogue reconstruction and discrepancy-aware reasoning.

\paragraph{Step 1: Asymmetric Interaction}
At iteration $k$, the user simulator $\pi_\theta$ generates the user utterance at turn $t$ as
\begin{equation}
\hat{u}^{(k)}_t \sim \pi_\theta(\cdot \mid \mathcal{P}_k, \hat{\mathcal{H}}^{(k)}_{<t}),
\end{equation}
where $\hat{\mathcal{H}}^{(k)}_{<t}$ is the simulated history before turn $t$. To stabilize the interaction environment, we adopt an asymmetric design on the assistant side: the assistant model is given the full original dialogue $\mathcal{D}$ as a global strategy reference, and is instructed to follow the original assistant's interaction strategy without copying the ground-truth responses verbatim. Its response is generated by
\begin{equation}
\hat{a}^{(k)}_t
=
\mathcal{M}_{\text{assist}}
\big(
\mathcal{I}_{\text{assist}}
\oplus
\mathcal{D}
\oplus
\hat{\mathcal{H}}^{(k)}_{<t}
\oplus
\hat{u}^{(k)}_t
\big).
\end{equation}
Rolling out for $T$ turns yields the simulated dialogue
\begin{equation}
\hat{\mathcal{D}}^{(k)}=\{(\hat{u}^{(k)}_t,\hat{a}^{(k)}_t)\}_{t=1}^{T}.
\end{equation}

\paragraph{Step 2: Chain-of-Thought Profile Optimization}
Given the real dialogue $\mathcal{D}$, the simulated dialogue $\hat{\mathcal{D}}^{(k)}$, and the current profile $\mathcal{P}_k$, a reasoning model $\mathcal{M}_{\text{reason}}$ analyzes their behavioral discrepancy and produces both a critique $\delta_k$ and an updated profile $\mathcal{P}_{k+1}$:
\begin{equation}
[\delta_k;\mathcal{P}_{k+1}]
=
\mathcal{M}_{\text{reason}}
\big(
\mathcal{I}_{\text{opt}}
\oplus
\mathcal{D}
\oplus
\hat{\mathcal{D}}^{(k)}
\oplus
\mathcal{P}_k
\big).
\end{equation}
This process is repeated for $N$ rounds, and the final optimized profile $\mathcal{P}^*=\mathcal{P}_N$ is used for downstream user simulation.

\subsection{Role-Reversal Supervised Fine-Tuning}
\label{sec:rrsft}
To align the simulator with authentic human behavior, we perform Supervised Fine-Tuning (SFT) using the optimized user profiles. We employ a Role-Reversal strategy: instead of predicting assistant responses from user instructions, we train the model to predict user utterances conditioned on the user profile and the full dialogue context. Specifically, for each dialogue $d_i$ and turn $j$, the model takes the IPSE-optimized profile $p_i$ together with the dialogue context $c_{i,j}$ as input, and predicts the corresponding user utterance $u_{i,j}$. The training objective is defined as:
\begin{equation}
\mathcal{L}_{\text{SFT}}(\theta)
=
-\sum_{i,j,k}
\log P_{\theta}\!\left(
u_{i,j,k}
\mid
u_{i,j,<k},\, c_{i,j},\, p_i
\right),
\end{equation}
where $u_{i,j,k}$ denotes the $k$-th token of the user utterance $u_{i,j}$, and $c_{i,j}$ denotes the full dialogue context prior to generating $u_{i,j}$. Consistent with standard SFT protocols, the loss is computed exclusively on the user response tokens, while the profile prompt and dialogue context are masked.

\subsection{Rubric-Guided Reinforcement Learning}
While Role-Reversal SFT equips the simulator with basic human-like patterns, it does not guarantee fine-grained behavioral alignment throughout extended multi-turn interactions. To address this limitation, we further optimize the user simulator through RL at the dialogue level.

Specifically, we formulate user simulation as a multi-turn reinforcement learning (RL) problem, where the policy $\pi_\theta$ is conditioned on the target user profile (Optimized via IPSE) and the evolving dialogue context, and iteratively generates user utterances over multiple turns. By interacting with the assistant model, the simulator rolls out a complete dialogue trajectory $d_i'$ under profile $p_i$.

The optimization objective is to maximize the expected reward assigned to the generated dialogue:
\begin{equation}
\max_{\theta}\;
\mathbb{E}_{p_i,\; d_i' \sim \pi_\theta(\cdot \mid p_i)}
\left[ R_\phi(d_i', p_i) \right],
\end{equation}
where $R_\phi$ denotes the rubric-based reward model that provides fine-grained supervision signals for simulated behavior. We first train this reward model to capture the target behavioral standards, and then use it to optimize the simulator policy in a multi-turn RL setting.

\subsubsection{Rubric-Based Reward Model Training}
\label{reward_model}
To optimize the simulator with reinforcement learning, we require a reward function that can provide scalable, stable, and fine-grained feedback at the utterance level. However, general-purpose LLM judges (e.g., GPT-4o) are often costly and unstable for rubric-based behavioral evaluation. Therefore, we train a specialized \textit{rubric-based reward model}.

Specifically, given a target profile $p_i$, a dialogue context $c_{i,j}$, and a candidate user utterance $u_{i,j}$, the reward model predicts an utterance-level score
\begin{equation}
\hat{s}_{i,j} = R_\phi(p_i, c_{i,j}, u_{i,j}),
\end{equation}
where $R_\phi$ denotes the reward model parameterized by $\phi$.

The reward-model training set is constructed by sampling profile-conditioned utterance instances from the collected training corpus, which are then annotated by experts with rubric-grounded rationales and scores.

\paragraph{Rubric Definition}
Our reward model evaluates each utterance against a rubric that covers three dimensions: (1) \textit{human likeness}, which judges whether the response sounds natural like a real human user in tone, slang, and sentence structure; (2) \textit{persona consistency}, which assesses whether the response follows the constraints set by the target profile; and (3) \textit{context coherence}, which examines whether the response is logically consistent with the dialogue history.

\paragraph{Training Strategy}
To learn this rubric effectively, we adopt a two-stage training strategy consisting of supervised warm-up followed by RLVR-based calibration.

\paragraph{Stage 1: Supervised Warm-up}
In the first stage, we perform SFT to teach the reward model to reproduce rubric-grounded evaluations, including expert-calibrated reasoning and target scores. This stage equips the model with an initial understanding of the scoring criteria and provides a strong starting point for subsequent refinement.

\paragraph{Stage 2: RLVR-based Calibration}
In the second stage, we further calibrate the reward model using Reinforcement Learning with Verifiable Rewards (RLVR), where the ground-truth score $s^*$ serves as the verifiable outcome. Since exact-match binary rewards are too coarse for our discrete scoring space $S=\{0, 0.5, 1\}$, we instead adopt a \textit{Distance-Aware Outcome Reward}:
\begin{equation}
R(\hat{s}, s^*) = 1 - |\hat{s} - s^*|,
\end{equation}
where $\hat{s}$ denotes the predicted score. This reward assigns partial credit to near-correct predictions, enabling finer-grained optimization of the model's scoring behavior than binary supervision.

The trained reward model provides turn-level reward signals for user simulation. In the subsequent multi-turn RL stage, these utterance-level rewards are aggregated into a session-level reward to optimize the simulator over complete dialogue trajectories.

\subsubsection{Multi-Turn RL via GRPO}
\label{sec:mtrl}

In this stage, the SFT-aligned user simulator interacts with the assistant model under a target profile $p_i$ to generate a simulated dialogue $d'_i$. Each dialogue consists of multiple user-assistant turns and terminates when reaching a predefined turn limit or the maximum context budget. 

For each generated user utterance $u'_{i,j}$ in the simulated dialogue $d'_i$, the trained reward model assigns an utterance-level reward based on the target profile and the current dialogue context:
\begin{equation}
r_{i,j} = R_\phi(p_i, c_{i,j}, u'_{i,j}),
\end{equation}
where $c_{i,j}$ denotes the dialogue context before generating the $j$-th user utterance.

Since our optimization objective is defined over complete interaction trajectories, we aggregate the turn-level rewards into a session-level reward by averaging across all user turns:
\begin{equation}
R_i = \frac{1}{T_i} \sum_{j=1}^{T_i} r_{i,j},
\end{equation}
where $T_i$ is the number of user turns in dialogue $d'_i$. We adopt Group Relative Policy Optimization (GRPO)~\cite{shao2024deepseekmathpushinglimitsmathematical} for RL training. 

By optimizing session-level rewards aggregated from rubric-based utterance-level evaluations, the simulator is encouraged to maintain consistent human-like behavior throughout the full interaction trajectory.

\section{Experiments}
\subsection{Dataset Construction}
To train and evaluate the performance of our proposed user simulator under diverse scenarios, we build a large-scale Chinese dialogue dataset collected from real human interactions. The dataset covers six distinct domains to fully embody various user personas, professional demands and personal preferences: General Chat (from WildChat \cite{zhao2024wildchat1mchatgptinteraction}), Customer Service (from JDDC \cite{chen-etal-2020-jddc}), Medical (from MedDialog \cite{he2020meddialoglargescalemedicaldialogue}), Legal (from Legal QA \cite{HanFei}), Sports and Entertainment (from NaturalConv \cite{Wang_2021}), and Technology and Education (also from NaturalConv). The detailed composition of our dataset is presented in Table \ref{tab:dataset_stats}, and further details are provided in the Appendix~\ref{sec:appendix-dataset}.

\begin{table}[t]
    \centering
    \resizebox{0.48\textwidth}{!}
    {
    \begin{tabular}{llr}
    \toprule
    \textbf{Domain} & \textbf{Source Dataset} & \textbf{Size (Sessions)} \\
    \midrule
    General Chat & WildChat & 2,572 \\
    Customer Service & JDDC & 1,293 \\
    Medical & MedDialog & 1,481 \\
    Legal & Legal QA & 1,476 \\
    Technology \& Education & NaturalConv & 1,472 \\
    Sports \& Entertainment & NaturalConv & 1,482 \\
    \midrule
    \textbf{Total} & - & \textbf{9,776} \\
    \bottomrule
    \end{tabular}
    }
    \caption{Statistics of the constructed User Simulation Dataset across six domains.}
    \label{tab:dataset_stats}
\end{table}

\paragraph{Preprocessing and Profile Annotation}
To ensure sufficient context for persona manifestation, we apply a strict filtering criterion and retain only dialogue sessions with at least four turns, where one turn consists of a user utterance followed by an assistant response. We then apply the proposed \textit{Iterative Profile Self-Evolution} (IPSE) framework to synthesize an optimized user profile $p_i$ for each dialogue session $d_i$, resulting in profile-conditioned training pairs $(p_i, d_i)$.

\paragraph{Difficulty-based SFT/RL Partition}
We split the corpus by dialogue difficulty for SFT and RL training. Each session is assigned a difficulty score based on {Constraint Density}, {Information Withholding}, and {Intent Volatility}, then ranked accordingly. The top 10\% most difficult sessions are reserved for RL, while the remainder are used for SFT. This allows SFT to focus on learning basic profile-conditioned patterns from lower-complexity interactions, while reserving highly constrained and strategically challenging dialogues for RL policy optimization.

\subsection{Implementation Details}
We use Qwen3-8B~\cite{yang2025qwen3technicalreport} as the backbone model in all experiments, and conduct all training stages in its non-reasoning mode. Role-Reversal Supervised Fine-Tuning is performed on the SFT subset with a maximum sequence length of 8192, for 3 epochs at a learning rate of $1\times10^{-6}$. 

The rubric-based reward model is trained in two stages. In the supervised warm-up stage, we fine-tune the model for 3 epochs with a learning rate of $5\times10^{-6}$. In the RLVR stage, we further optimize it for 5 epochs with a learning rate of $1\times10^{-6}$.

For multi-turn RL, we initialize the simulator from the SFT-aligned policy and optimize it with GRPO on the RL subset. The group size is set to 4, and training is conducted for 5 epochs with a learning rate of $1\times10^{-6}$.

\begin{table*}[t]
\renewcommand{\arraystretch}{1.2}
\centering
\small
\setlength{\tabcolsep}{5pt}
\caption{Main results on utterance-level and session-level evaluation. Best results are highlighted in \textbf{bold}.}
\label{tab:main_results}

\begin{tabular*}{\textwidth}{@{\extracolsep{\fill}}lccccccc@{}}
\toprule
\multicolumn{8}{c}{\textbf{Panel A: Utterance-Level Evaluation}} \\
\midrule
\multirow{2}{*}{\textbf{Model}} 
& \multicolumn{3}{c}{\textbf{Objective Metrics}} 
& \multicolumn{4}{c}{\textbf{LLM-Judge Metrics}} \\
\cmidrule(lr){2-4} \cmidrule(lr){5-8}
& \textbf{AI Prob.}$\downarrow$
& \textbf{Style Sim.}$\uparrow$
& \textbf{AVA}$\uparrow$
& \textbf{Contxt. Relv.}$\uparrow$
& \textbf{Resp. Fid.}$\uparrow$
& \textbf{Goal Contr.}$\uparrow$
& \textbf{Ling. Nat.}$\uparrow$ \\
\midrule
UserLM        & 69.92 & 0.5888 & 55.38 & 0.5111 & 0.5289 & 0.4742 & 0.5655 \\
USP           & \underline{43.14} & 0.6644 & 59.25 & 0.6384 & 0.6874 & 0.5779 & 0.7314 \\
Qwen3-8B      & 44.45 & \underline{0.7307} & \underline{64.69} & 0.8644 & 0.9102 & 0.7888 & 0.8976 \\
GPT-4o        & 45.50 & 0.7181 & 62.79 & \underline{0.8891} & \underline{0.9221} & \underline{0.8304} & \underline{0.9214} \\
\midrule
\rowcolor{gray!8}
Muse (w/o RL) & 37.98 & \textbf{0.7606} & 64.80 & 0.9046 & 0.9591 & 0.8530 & 0.9389 \\
\rowcolor{blue!6}
\textbf{Muse} & \textbf{31.18} & 0.7534 & \textbf{64.89} & \textbf{0.9196} & \textbf{0.9776} & \textbf{0.8763} & \textbf{0.9620} \\
\bottomrule
\end{tabular*}

\vspace{0.45em}

\begin{tabular*}{\textwidth}{@{\extracolsep{\fill}}lccccc@{}}
\toprule
\multicolumn{6}{c}{\textbf{Panel B: Session-Level Evaluation}} \\
\midrule
\textbf{Model}
& \textbf{Pers. Cons.}
& \textbf{Goal. Eff.}
& \textbf{Dial. Coh.}
& \textbf{Constr. Comp.}
& \textbf{Avg. Score} \\
\midrule
UserLM        & 0.3284 & 0.5208 & 0.6336 & 0.2942 & 0.4442 \\
USP           & 0.4266 & 0.5882 & 0.7692 & 0.4038 & 0.5469 \\
Qwen3-8B      & 0.7415 & 0.6256 & 0.7159 & 0.7181 & 0.7003 \\
GPT-4o        & \underline{0.7625} & \underline{0.7586} & \underline{0.8187} & \underline{0.7834} & \underline{0.7808} \\
\midrule
\rowcolor{gray!8}
Muse (w/o RL) & 0.6813 & 0.6414 & 0.7273 & 0.6825 & 0.6831 \\
\rowcolor{blue!6}
\textbf{Muse} & \textbf{0.8378} & \textbf{0.7967} & \textbf{0.8685} & \textbf{0.8739} & \textbf{0.8442} \\
\bottomrule
\end{tabular*}
\end{table*}

\subsection{Evaluation Metrics}
\label{sec:metrics}

To comprehensively assess the effectiveness of \textbf{MUSE} in user simulation, we conduct evaluations at two granularities: utterance-level metrics for single-turn quality assessment and session-level metrics for multi-turn holistic evaluation.

\subsubsection{Utterance-Level Metrics}

At the utterance level, we employ both objective automatic metrics and LLM-as-a-Judge criteria to evaluate the quality of individual generated responses.

\paragraph{Objective Metrics.} We adopt three objective metrics to assess stylistic authenticity: (1) \textbf{AI-generated Probability (AI Prob)}, where we utilize a binary AI detection model~\cite{wang2025knowbettermodelinghumanlike} to estimate the probability that a generated utterance is AI-generated, with lower values indicating more human-like text; (2) \textbf{Style Similarity (Style Sim)}, where we employ a style embedding model~\cite{wegmann-etal-2022-author} to measure the stylistic consistency between the generated response and the reference user utterance; and (3) \textbf{Author Verification Accuracy (AVA)}, derived from~\cite{wegmann-etal-2022-author,wang2025knowbettermodelinghumanlike}, which evaluates whether sentence pairs are attributed to the same author based on similarity thresholds, capturing content-independent style representations.

\paragraph{LLM-as-a-Judge Metrics.} To assess dialogue quality beyond semantic and stylistic perspectives, we leverage DeepSeek-V3~\cite{deepseekai2025deepseekv3technicalreport} as an evaluator to score four critical dimensions (0–1 scale)): (1) \textbf{Contextual Relevance}, measuring whether the current response precisely addresses the dialogue history without topic drift or mechanical repetition; (2) \textbf{Response Fidelity}, evaluating strict adherence to persona constraints and role-appropriate behavior; (3) \textbf{Goal Contribution}, assessing the contribution of the current turn toward the conversational objective; and (4) \textbf{Linguistic Naturalness}, judging whether the text conforms to human conversational habits with semantic fluency and logical coherence.

\subsubsection{Session-Level Metrics}

While utterance-level metrics capture local quality, they fail to fully reflect the simulator's ability to maintain consistent personas across extended interactions. Therefore, we use an assistant model to conduct multi-turn dialogues with the user model, introducing four session-level metrics: (1) \textbf{Persona Consistency}, measuring stable role perspectives and emotional tendencies throughout the conversation; 
(2) \textbf{Goal-oriented Effectiveness}, assessing task advancement efficiency and strategic adjustments; 
(3) \textbf{Dialogue Coherence}, evaluating natural topic transitions and information consistency across turns; and (4) \textbf{Constraint Compliance}, evaluating adherence to hard constraints across the full session. These metrics are averaged for an overall session-level assessment.

\begin{table*}[t]
\centering
\small
\renewcommand{\arraystretch}{1.25}
\setlength{\tabcolsep}{6pt}
\caption{Ablation results of different reward model training strategies. Best results are highlighted in \textbf{bold}.}
\label{tab:rm_ablation}
\begin{tabular*}{\textwidth}{@{\extracolsep{\fill}}lcccccccc@{}}
\toprule
\multirow{2}{*}{\textbf{Method}} 
& \multicolumn{2}{c}{\textbf{Human Likeness}} 
& \multicolumn{2}{c}{\textbf{Persona Consistency}} 
& \multicolumn{2}{c}{\textbf{Context Coherence}} 
& \multicolumn{2}{c}{\textbf{Overall}} \\
\cmidrule(lr){2-3} \cmidrule(lr){4-5} \cmidrule(lr){6-7} \cmidrule(lr){8-9}
& \textbf{Acc} & \textbf{EM} 
& \textbf{Acc} & \textbf{EM} 
& \textbf{Acc} & \textbf{EM} 
& \textbf{Acc} & \textbf{EM} \\
\midrule
Qwen3-8B  & 0.7087 & 0.4400 & 0.6875 & 0.4250 & 0.6725 & 0.4375 & 0.6896 & 0.4342 \\
Qwen3-8B-SFT w/o CoT & 0.9087 & 0.8200 & 0.8525 & 0.7550 & 0.8712 & 0.8050 & 0.8775 & 0.7933 \\
Qwen3-8B-SFT w/ CoT & 0.9125 & 0.8325 & 0.8725 & 0.7850 & 0.8888 & 0.8300 & 0.8912 & 0.8158 \\
\midrule
Qwen3-8B-RL w/o CoT & 0.8975 & 0.8175 & 0.8612 & 0.7525 & 0.9012 & 0.8550 & 0.8867 & 0.8083 \\
\rowcolor{blue!6}
Qwen3-8B-RL w/ CoT & \textbf{0.9213} & \textbf{0.8425} & \textbf{0.8862} & \textbf{0.7950} & \textbf{0.9126} & \textbf{0.8600} & \textbf{0.9067} & \textbf{0.8333} \\
\bottomrule
\end{tabular*}
\end{table*}

\begin{table}[t]
\renewcommand{\arraystretch}{1.2}
    \centering
    \resizebox{0.48\textwidth}{!}
    {
    \begin{tabular}{l l c}
        \toprule
        Method & Extraction Model & PPL $\downarrow$ \\
        \midrule
        Single-pass Profile Extraction
 & Gemini-3-Flash & 85.97 \\
        Single-pass Profile Extraction
 & DeepSeek-V3 & 77.86 \\
 \midrule
    \rowcolor{blue!6} \textbf{IPSE Framework (Ours)} & \textbf{DeepSeek-V3} & \textbf{73.12} \\
        \bottomrule
    \end{tabular}
    }
    \caption{Perplexity (PPL) of different profile extraction methods on the test set.}
    \label{tab:ab-ipse}
\end{table}

\subsection{Results}
\label{sec:results}

\subsubsection{Utterance-Level Evaluation}

Table~\ref{tab:main_results} (Panel A) reports the main results at the utterance level. We observe that UserLM~\cite{naous2026flippingdialoguetrainingevaluating} and USP~\cite{wang2025knowbettermodelinghumanlike} remain clearly behind strong general-purpose baselines in both objective metrics and LLM-judge metrics. Qwen3-8B and GPT-4o achieve substantially better local response quality, indicating that stronger base models already provide advantages in relevance, response fidelity, and naturalness. Compared with these strong baselines, MUSE further improves utterance-level performance and achieves the best results on nearly all metrics. Moreover, the full MUSE model improves most utterance-level metrics over \textit{Muse (w/o RL)}, suggesting that the RL stage provides additional gains beyond supervised imitation alone.

\subsubsection{Session-Level Evaluation}
At the session level, the advantage of MUSE becomes more pronounced, as shown in Table~\ref{tab:main_results} (Panel B). While strong baselines like GPT-4o remain competitive in extended interactions, MUSE consistently surpasses them across all session-level metrics. In contrast, although \textit{Muse (w/o RL)} remains competitive, its gains over strong baselines are limited at the dialogue level. This comparison shows that strong utterance‑level generation alone cannot ensure stable long‑horizon behavioral alignment. By further introducing rubric-guided multi-turn RL, MUSE obtains substantial gains across all session-level dimensions. These improvements reflect stronger persona maintenance, more coherent dialogue progression, and better adherence to conversational constraints throughout the interaction.

\section{Ablation Study}

\subsection{Effectiveness of IPSE}

To evaluate our Iterative Profile Self-Evolution (IPSE) framework, we use perplexity (PPL) to measure how well the language model fits the data distribution under different user profiles. As shown in Table~\ref{tab:ab-ipse}, IPSE reduces PPL from 77.86 to 73.12, demonstrating that iterative refinement yields more accurate profiles than single-pass extraction with DeepSeekV3 or Gemini-3-Flash, which consistently produce higher PPL scores due to insufficient capture of user preferences.

\subsection{Ablations on Reward Model Training}
Table~\ref{tab:rm_ablation} presents ablation studies on reward model training, including the efficacy of the progressive SFT-RLVR pipeline and the specific contribution of Chain-of-Thought (CoT) reasoning. We report Accuracy (Acc) and Exact Match Rate(EM) to quantify the alignment with expert annotations.

\paragraph{Effectiveness of the SFT-RLVR Pipeline}
We assess the gains from our two-stage training. As shown in Table~\ref{tab:rm_ablation}, the vanilla Qwen3-8B baseline achieves only 0.6896 Overall Acc, indicating off-the-shelf models struggle with our rubric nuances. Applying SFT yields a substantial improvement, boosting Acc to 0.8912. Furthermore, the introduction of RLVR provides the final leap in performance. By optimizing the distance-aware outcome reward ($1 - |\hat{s} - s^*|$), the model achieves a peak Overall Acc of {0.9067} and an EM of {0.8333}. This trajectory demonstrates that while SFT instills the basic evaluation format, the RLVR stage is indispensable for strictly aligning the model's scoring logic with expert standards.

\paragraph{Critical Role of Chain-of-Thought}
We further analyze CoT's impact across training stages. In the SFT phase, integrating CoT outperforms the score-only approach (0.8912 vs. 0.8775), serving as a necessary cognitive scaffold. Crucially, this advantage is amplified in the RL stage, where optimizing the full reasoning-to-score path significantly surpasses configurations without CoT. This validates that reasoning traces are essential for maximizing the reward model's effectiveness.

\subsection{Ablations on the Training Pipeline}

We conduct ablation studies to assess each stage's contribution, following the optimization path \textit{Qwen3-8B} $\rightarrow$ \textit{Muse (w/o RL)} $\rightarrow$ \textit{Muse}. As shown in Table~\ref{tab:main_results}, Role-Reversal SFT consistently improves utterance-level performance, indicating that supervised adaptation effectively transfers the base model to profile-conditioned user simulation. However, SFT achieves similar session-level performance to the baseline, as it primarily learns stylistic patterns without improving long-term role retention.
Building on this, rubric-guided multi-turn RL further improves both utterance-level and session-level results, with especially large gains in persona consistency, constraint compliance, and overall dialogue quality. These results confirm that SFT strengthens local response quality, while RL is critical for maintaining robust behavioral alignment over extended interactions.

\section{Conclusion}
In this paper, we present \textbf{MUSE} (\textbf{M}ulti-Domain Chinese \textbf{U}ser \textbf{S}imulation via self-\textbf{E}volving Profiles and Rubric-Guided Alignment), a user simulation framework designed to generate high-fidelity and consistent user behaviors. Through \textit{Iterative Profile Self-Evolution} (IPSE), MUSE synthesizes diverse and persona-consistent user profiles that provide a strong foundation for controllable simulation. Building on this, \textit{Role-Reversal SFT} aligns the simulator with realistic human response patterns, while the proposed \textit{rubric-based reward model} and \textit{rubric-guided multi-turn RL} further enforce fine-grained behavioral alignment throughout extended interactions. Experimental results show that MUSE consistently outperforms strong baselines at both the utterance and session levels, demonstrating its effectiveness as a robust, scalable, and controllable solution for user simulation in agent training, synthetic data generation, and multi-turn evaluation.

\clearpage

\newpage
\section*{Limitations}
Although MUSE achieves strong performance in multi-domain Chinese user simulation, several limitations remain. First, our study is conducted on Chinese dialogue data from six domains, and the extracted profiles, behavioral rubrics, and learned interaction patterns may therefore encode language- and domain-specific priors. As a result, the generalizability of MUSE to other languages, cultures, and more specialized or unseen domains remains unclear. Moreover, the current RL setting is still bounded by predefined turn limits and context budgets, so the behavior of MUSE in substantially longer or more open-ended interactions remains underexplored. We leave cross-lingual transfer and more efficient long-horizon simulation to future work.

\section*{Ethical Statement}
This work uses existing Chinese dialogue datasets from six domains to build and evaluate a multi-domain user simulator. Because these data may contain sensitive, offensive, or potentially identifiable content, we rely on previously released datasets, apply preprocessing before profile construction, and use extracted profiles only as abstract behavioral descriptions rather than real-user identifiers. We also acknowledge possible bias from expert annotations and LLM-based evaluation, as well as the potential misuse of user simulators to imitate human behavior. AI assistants were used only to improve writing clarity, while all research design, experiments, and final verification were carried out by the authors.

\bibliography{main}

\newpage
\appendix

\section{Case Study of IPSE}
\label{sec:appendix-ipse}

To demonstrate how IPSE reconstructs complex user simulators from raw data, we present a case study comparing the initial extraction against the final profile evolved via IPSE. The source data is a real-world multi-turn conversation between a lawyer and an assistant.

\subsection{Ground Truth: Raw Dialogue}
\label{sec:raw_data}
The following transcript represents the raw interaction history used as the seed data.
\textit{Key Observation: The user employs a specific strategy—starting with broad inquiries, engaging in critique, raising pain points (``freeloading''), and withholding the specific success case until the final turn.}

\begin{quote}
\small
\textbf{User:} Hello. I am a Chinese lawyer. Can you speak Chinese? \\
\textbf{Assistant:} Hello! Sure. \\
\textbf{User:} I want to do short video promotion. Do you have any scripts? \\
\textbf{Assistant:} [Provides generic scripts: ``Everyone needs legal help''...] \\
\textbf{User:} \textbf{I feel that the above scripts have no attraction.} Can you analyze it from a self-media perspective? \\
\textbf{Assistant:} [Provides improved scripts: ``Legal Tips''...] \\
\textbf{User:} As a lawyer, many clients consult but don't pay. \textbf{How can I avoid them ``freeloading''?} \\
\textbf{Assistant:} [Suggests signing contracts and optimizing flow...] \\
\textbf{User:} \textbf{Actually, here is a specific case:} I guided a friend to recover \textbf{20,000 RMB wages} through arbitration... It took a year. I want a script based on this.
\end{quote}

\subsection{Evolution of Extracted Profiles}

We compare the baseline profile (Iteration 0) with the final structured profile (Iteration 2) to highlight the impact of IPSE.

\begin{tcolorbox}[colback=gray!10, colframe=gray!50, title=\textbf{[Profile Iteration 0: Baseline Extraction]}]
\small
\textit{Deficiency: Information Flattening.} The profile captures the facts but flattens the temporal logic. Crucially, the specific ``20k wage case'' is listed as a static background attribute. A simulator using this profile would likely ``leak'' this information in the very first turn, violating the ground truth flow.

\vspace{0.5em}
\textbf{Background:} You are a Chinese lawyer trying to promote via Douyin. \\
\textbf{Goal:} Get eye-catching scripts to increase entrustment rates. \\
\textbf{Static Context:}
\begin{itemize}
    \setlength\itemsep{0em}
    \item You successfully helped a friend recover \textbf{20,000 RMB wages} via labor arbitration and want to make a video about it. \textcolor{red}{\textbf{[Error: Marked as immediate background]}}
\end{itemize}
\textbf{Style:} Direct, pragmatic, focuses on conversion.
\end{tcolorbox}

\vspace{1em} 

\begin{tcolorbox}[colback=green!5, colframe=green!40, title=\textbf{[Profile Iteration 2: Final IPSE Construction]}]
\small
\textit{Success: Structured Cognitive Roadmap.} The profile is transformed into a structured strategy. It explicitly categorizes the ``20k case'' as a \textbf{Hidden Context} and defines a 5-stage conversation flow, perfectly reconstructing the user's implicit strategy.

\vspace{0.5em}
\textbf{Role:} User Simulator - Chinese Lawyer. \\
\textbf{Core Task:} Test a short-video marketing AI. \\
\textbf{Persona (Attributes):}
\begin{itemize}
    \setlength\itemsep{0em}
    \item \textbf{Pain Points:} Knowledgeable in law but new to video; frustrated by ``freeloaders''.
    \item \textbf{Personality:} Concise, results-oriented. If the AI is too generic, \textbf{criticize it directly}.
    \item \textbf{Hidden Context:} A real success case (recovering 20k RMB wages). \textcolor{blue}{\textbf{[Note: Do not reveal initially. Wait for the right moment.]}}
\end{itemize}

\textbf{Conversation Flow (Structured Strategy):}
\begin{enumerate}
    \setlength\itemsep{0em}
    \item \textbf{Stage 1 (Ice-breaking):} State identity: ``I'm a lawyer...''
    \item \textbf{Stage 2 (Critical Thinking):} If the script is generic, \textbf{you must criticize it} (e.g., ``No attraction'').
    \item \textbf{Stage 3 (Business Pain Points):} Ask about monetization: ``How to avoid free-riders?''
    \item \textbf{Stage 4 (Specific Context Injection):} \textit{Trigger: Only after previous stages.} \textbf{Reveal the ``20k RMB wage case''} and ask for a plot twist script.
    \item \textbf{Stage 5 (Closing):} Express satisfaction.
\end{enumerate}
\end{tcolorbox}

\subsection{Analysis}
The direct comparison between Iteration 0 and Iteration 2 highlights the core contribution of IPSE. While the baseline extraction merely summarizes \textit{what} the user knows, IPSE successfully infers \textit{when} and \textit{how} the user intends to share that information. By structuring the ``20k case'' into \textbf{Stage 4} of the conversation flow, IPSE ensures the simulator replicates the authentic, multi-turn trust-building process observed in the raw data, preventing the common issue of information leakage in user simulation.

\section{Details of the Used Datasets}
\label{sec:appendix-dataset}

We provides a detailed description of the Chinese dialogue dataset constructed to train and evaluate our user simulator. We elaborate on the data source, composition, and characteristics of each domain, as follows:

\begin{itemize}
    \item \textbf{General Chat}: Sourced from \textbf{WildChat} \cite{zhao2024wildchat1mchatgptinteraction}, this domain contains the largest volume of data. Its high diversity significantly enhances the model's capability to simulate a wide range of generic user personas in open-ended conversations.
    \item \textbf{Customer Service}: Derived from the \textbf{JDDC} dataset~\cite{chen-etal-2020-jddc}, focusing on goal-oriented interactions between customers and service agents.
    \item \textbf{Medical}: Sourced from \textbf{MedDialog}~\cite{he2020meddialoglargescalemedicaldialogue}, covering doctor-patient consultations and health-related inquiries.
    \item \textbf{Legal}: A dedicated multi-turn \textbf{Legal QA}~\cite{HanFei} dataset, representing scenarios where users seek professional legal advice.
    \item \textbf{Sports \& Entertainment}: Sourced from the \textbf{NaturalConv}~\cite{Wang_2021} dataset, this domain captures interest-driven discussions regarding sports events, celebrities, and entertainment news.
    \item \textbf{Technology \& Education}: Also sourced from \textbf{NaturalConv}, this subset focuses on knowledge-sharing interactions related to tech trends, educational topics, and scientific discussions.
\end{itemize}

\end{document}